\pdfoutput=1

\documentclass[11pt]{article}

\usepackage[]{acl}

\usepackage{times}
\usepackage{latexsym}

\usepackage[T1]{fontenc}

\usepackage[utf8]{inputenc}

\usepackage{microtype}
\usepackage{booktabs}
\usepackage{multirow}
\usepackage{scalerel}
\usepackage{graphicx}
\usepackage{changepage}  
\usepackage{subcaption}
\usepackage[compact]{titlesec}
\usepackage{comment}

\usepackage{tabularx}

\newcommand{\al}[1]{\textcolor{black}{#1}}

\newcommand{\fire}[0]{\scalerel*{\includegraphics{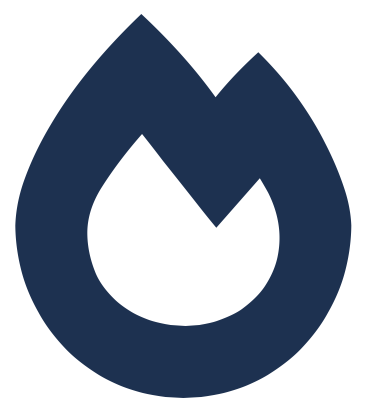}}{\strut}}
\title{Bridging Fairness and Environmental Sustainability \\in Natural Language Processing}

  \author{Marius Hessenthaler\textsuperscript{1}, Emma Strubell\textsuperscript{2}, Dirk Hovy\textsuperscript{3}, Anne Lauscher\textsuperscript{4} \\
  \textsuperscript{1}Data and Web Science Group, University of Mannheim, Germany \\
  \textsuperscript{2}Language Technologies Institute,  Carnegie Mellon University, U.S.  \\
    \textsuperscript{3}MilaNLP,  Bocconi University, Italy \\
        \textsuperscript{4}Data Science Group,  University of Hamburg, Germany \\
  \small{\texttt{marius.hessenthaler@web.de}, \texttt{strubell@cmu.edu}},\\ \small{\texttt{dirk.hovy@unibocconi.it}, \texttt{anne.lauscher@uni-hamburg.de}} \\}

\begin{document}
\maketitle
\begin{abstract}
\emph{Fairness} and \emph{environmental impact} are important research directions for the sustainable development of artificial intelligence. However, while each topic is an active research area in natural language processing (NLP), there is a surprising lack of research on the interplay between the two fields. 
This lacuna is highly problematic, since there is increasing evidence that an exclusive focus on fairness can actually hinder environmental sustainability, and vice versa. 
In this work, we shed light on this crucial intersection in NLP by (1) investigating the efficiency of current fairness approaches through surveying example methods \al{for reducing unfair stereotypical bias} from the literature, and (2) evaluating a common technique to reduce energy consumption (and thus environmental impact) of \al{English} NLP models, knowledge distillation (KD), for its impact on fairness. 
In this case study, we evaluate the effect of important KD factors, including layer and dimensionality reduction, with respect to: (a)~performance on the distillation task (natural language inference and semantic similarity prediction), and (b)~multiple measures and dimensions of stereotypical bias (e.g., gender bias measured via the Word Embedding Association Test). 
Our results lead us to \al{clarify current assumptions} regarding the effect of KD on unfair bias: contrary to \al{other} findings, we show that KD can actually \textit{decrease} model fairness. 
\end{abstract}

\section{Introduction}
Fairness and environmental sustainability are critical to the future of human society, and, thus, also reflected by the United Nations' 17 Sustainable Development Goals (e.g., \emph{Goal 5: Gender Equality}, and \emph{Goal 13: Climate Action}).\footnote{\url{https://sdgs.un.org/goals}}
Accordingly, both topics are currently also  active research areas in natural language processing (NLP). 

On the one hand, several works have established that language representations are prone to encode and amplify stereotypical social biases~\cite[e.g.,][]{Bolukbasi}, and, consequently, are a source of representational harm~\citep{barocas2017problem,hovy-spruit-2016-social,shah-etal-2020-predictive}. To address this issue and provide fairer language technologies, various approaches have developed methods for measuring bias~\citep[e.g.,][\emph{inter alia}]{Caliskan_2017, nadeem2020stereoset, nangia-etal-2020-crows,nozza-etal-2021-honest} as well as debiasing methods~\citep[e.g.,][\emph{inter alia}]{zhao-etal-2018-gender,dev2019attenuating}. %

On the other hand, recent advances in NLP have been fueled largely by increasingly computationally expensive pre-trained language models (PLMs). Whereas the original BERT base model has 110M parameters~\citep{devlin-etal-2019-bert}, the Switch Transformer model, designed as a more efficient alternative to more recent PLMs, has over a trillion parameters~\citep{JMLR:v23:21-0998}. 
While these models consistently obtain superior performance across a variety of NLP benchmarks~\citep{wang-etal-2018-glue,NEURIPS2019_4496bf24}, researchers have pointed out the increasing %
potential CO$_2$ emissions of these models. \citet{strubell-etal-2019-energy} estimated that pre-training a BERT base Transformer~\citep{Transformer} using energy with average U.S. carbon intensity has CO$_2$ emissions comparable to a passenger on a trans-American flight. More recent calculations confirm that the energy consumption of PLMs continues to grow along with their size~\citep{dodge2022measuring} and that consumption at inference time is non-negligible~\citep{10.1145/3466752.3480095}. %
These findings have fueled the development of more environmentally sustainable NLP. 
For instance, tuning only a few new lightweight adapter layers instead of the whole %
architecture~\citep[e.g.,][]{Adapters,pfeiffer2021adapterfusion}, and compressing models~\citep[e.g.,][]{10.1145/3487045} can reduce the energy consumption during training or inference.

However, while both fairness and sustainability are active research fields in our community,\footnote{See also the proceedings of dedicated workshops, e.g., SustaiNLP (\url{https://aclanthology.org/2021.sustainlp-1.0/}, and LT-EDI (\url{https://aclanthology.org/2022.ltedi-1.0/}))} it is extremely surprising that there is so little work on the intersection of \textit{both} aspects. We argue that \textbf{this lack of focus is problematic, as some fairness approaches can  jeopardize sustainability and sustainability approaches might hinder fairness}.

For instance, \citet{webster2020measuring} propose a data-driven debiasing approach,  %
which requires pre-training a fairer model from scratch. Thus, for each and every stereotype, a novel PLM must be trained, reducing environmental sustainability. \citet{lauscher2021sustainable} pointed to potential issues of such an approach and proposed a modular, and therefore more sustainable method. In the other direction, recent work in %
computer vision has shown that compressed models are less robust, %
and can even amplify algorithmic bias %
\citep{hooker2020characterising,liebenwein2021lost}. \citet{ahia-etal-2021-low-resource} investigated the relationship between %
pruning and low-resource machine translation, finding that pruning can actually aid generalization in this scenario by reducing undesirable memorization. However, aside from few works, %
there has been no systematic research on the interplay between the two fields in NLP.

\paragraph{Contributions.} 
In this work, we acknowledge the potential for race conditions between fairness and environmental sustainability in NLP and call for more research on the interplay between the two fields. To shed light on the problem and to provide a starting point for fair and environmentally sustainable NLP,
\textbf{(1)} we \al{provide a literature overview} and systematize \al{a selection of} exemplary fairness approaches according to their sustainability aspects. We show that the surveyed approaches require energy at various training stages and argue that fairness research should consider these aspects. 
\textbf{(2)} Based on work suggesting the potential of model compression to increase fairness~\citep{xu2022can}, we take a closer look at  knowledge distillation~\citep[KD;][]{hinton2015distilling} as an example method  targeting the environmental sustainability of language technology. In this approach, a (smaller) student model is %
guided by the knowledge of a (bigger) teacher model. We extensively analyze the effect of KD on intrinsic and extrinsic bias measures (e.g., Word Embedding Association Test~\citep[e.g.,][]{Caliskan_2017}, Bias-NLI~\citep{dev2020measuring}) across two tasks (Natural Language Inference and Semantic Similarity Prediction). We  investigate important KD-factors, such as the number of hidden layers of the student and their dimensionality. Contrary to concurrent findings~\citep{xu2022can}, we show that KD can actually \textit{decrease} fairness. Thus, fairness in such sustainability approaches needs to be carefully monitored. %
\al{We hope to inspire and inform future research into fair and environmentally sustainable language technology and make all code produced publicly available at: \url{https://github.com/UhhDS/knowledge_distillation_fairness}.}

\section{How Fairness Can Harm Sustainability}
To illustrate the tight relationship between environmental sustainability and fairness in current NLP, we conduct an exemplary analysis of current mitigation approaches for unfair bias. Here, our goal is not to conduct an exhaustive survey, but to showcase \emph{when, why, and to what extent} fairness approaches can be environmentally harmful. 

\subsection{Approach} 
We query the ACL Anthology\footnote{\url{https://aclanthology.org}} for \emph{``debiasing''} and \emph{``bias mitigation''} and examine the first 20 results each. We focus on debiasing of unfair societal stereotypes in monolingual PLMs. Therefore, we exclude approaches on static embeddings, domain generalization,\footnote{As for instance common in the fact verification literature~\citep[e.g.,][]{paul-panenghat-etal-2020-towards}} and solely multilingual PLMs. We also consider only papers that propose a novel adaptation or debiasing approach, and exclude papers that survey or benchmark mitigation methods~\citep[e.g.,][]{meade-etal-2022-empirical}. We remove any duplicates. 

This approach left us with 8 relevant publications (out of the initial 40 ACL Anthology hits). To diversify the analysis pool, we added one more paper, based on our expert knowledge.

If a paper proposes multiple methods, we focus only on a single method. We apply a coarse-grained distinction between (a)~\emph{projection-based}, and (b)~\emph{training-based} methods. Projection-based methods follow an analytical approach in a manner similar to the classic hard debiasing~\citep{Bolukbasi}. %
In contrast, training-based methods either rely on augmenting training sets~\citep[e.g.,][]{zhao-etal-2018-gender} or on %
a dedicated 
debiasing loss~\citep[e.g.,][]{qian-etal-2019-reducing}. %
For the training-based approaches, we additionally classify the stage where the authors demonstrate the debiasing.

\setlength{\tabcolsep}{3pt}
\begin{table*}[!t]\centering

\small
\begin{tabular}{lllcccccc}\toprule
\textbf{} & &\textbf{} &\multicolumn{5}{c}{\textbf{Increased Environmental Costs?}} &\textbf{} \\\cmidrule{4-8}
\textbf{Reference} &\textbf{Method} &\textbf{Type} &\textbf{0. Pre-t.} &\textbf{1. Inter.} &\textbf{2. Fine-t.} &\textbf{3. Inf.} &\textbf{Other} \\\midrule
\citet{karve-etal-2019-conceptor} & \emph{Conceptor Debiasing} &Projection & {\color{lightgray}--}  & {\color{lightgray}--} & {\color{lightgray}--} & {\color{lightgray}--} & \fire\  \\
\citet{liang-etal-2020-towards} & \emph{Sent-Debias} & Projection & {\color{lightgray}--}  & {\color{lightgray}--} & {\color{lightgray}--} & {\color{lightgray}--} & \fire\  \\
\citet{kaneko-bollegala-2021-debiasing} & \emph{Debias Context. Embs.} & Project. \& Train. &{\color{lightgray}--} &{\color{lightgray}--} &\fire\  &{\color{lightgray}--} & {\color{lightgray}--}\\
\citet{webster2020measuring} & \emph{Pre-training CDA} &Training &\fire\ \fire\ \fire\   & {\color{lightgray}--}&{\color{lightgray}--} &{\color{lightgray}--} & {\color{lightgray}--}\\
\citet{barikeri-etal-2021-redditbias} & \emph{Attribute Distance Deb.} &Training &{\color{lightgray}--} & \fire\ \fire\   &{\color{lightgray}--} & {\color{lightgray}--}& {\color{lightgray}--}\\
\citet{guo-etal-2022-auto} & \emph{Auto-Debias} & Training & {\color{lightgray}--}\ & \fire\ \fire\   & {\color{lightgray}--}& {\color{lightgray}--}\ & \fire\  \\
\citet{dinan-etal-2020-queens} & \emph{Biased-controlled Training} &Training & {\color{lightgray}--} & \fire\ \fire\  &{\color{lightgray}--} &{\color{lightgray}--} &{\color{lightgray}--} \\
\citet{subramanian-etal-2021-evaluating} & \emph{Bias-constrained Model} &Training & {\color{lightgray}--}&{\color{lightgray}--}&\fire\  & {\color{lightgray}--}& {\color{lightgray}--}\\

\citet{lauscher2021sustainable} & \emph{Debiasing Adapters} &Training&{\color{lightgray}--} &\fire\  &(\fire) &\fire\  & {\color{lightgray}--}\\
\bottomrule
\end{tabular}
\caption{Overview of examplary debiasing methods  w.r.t. their efficiency. We provide information on the type of the approach (\emph{Projection} vs. \emph{Training}), and estimate their environmental impact in 3 classes (\fire-- \fire \fire\fire) in different stages of the NLP-pipeline: \emph{0. Pre-training}, \emph{1. Intermediate Training}, \emph{2. Fine-tuning}, \emph{3. Inference time}, and \emph{Other}.}\label{tab:papers}
\end{table*}

\subsection{Results and Discussion} 
We show the results of our analysis in Table~\ref{tab:papers}.

\paragraph{Underlying Debiasing Approach.} Our small survey yielded examples from a variety of approaches: the \emph{projection-based} approaches are represented by \citep{karve-etal-2019-conceptor},  \citep{liang-etal-2020-towards}, and \citep{kaneko-bollegala-2021-debiasing}. These require generally only a small amount of energy~(\fire) for the analytical computation, which, in some cases, is iteratively applied to improve debiasing performance~\citep{ravfogel-etal-2020-null}. In this case, each iteration will marginally decrease the efficiency. \citet{kaneko-bollegala-2021-debiasing} explicitly couple their approach with the model fine-tuning.  In contrast, the other 6 works belong to the category of training-based approaches. Here, \citet{webster2020measuring} and \citet{lauscher2021sustainable} rely on CDA~\citep{zhao-etal-2018-gender} and \citet{dinan-etal-2020-queens} use control codes to guide the biases. \citet{barikeri-etal-2021-redditbias} rely on a loss-based bias mitigation for equalizing the distance of opposing identity terms towards sterotypical attributes. \citet{subramanian-etal-2021-evaluating} use a two-player zero-sum game approach for enforcing fairness constraints and \citet{guo-etal-2022-auto} rely on a prompt-based approach.

\paragraph{Training Stage.} 
For the projection-based approaches, the point in time of their application is not critical to their energy consumption. They can only be applied on a trained model (stages 1--3) and, in general, do not require much energy. 

However, for the training-based approaches, the training stage is a vital factor: using them in pre-training (stage 0) corresponds to training a new model from scratch. The energy (and corresponding CO$_2$ emissions) to perform full PLM pretraining can vary widely. Recent estimates range from 37.3 kWh to train BERT small, to  103.5 MWh to train a 6B parameter Transformer language model (\fire\fire\fire) \citep{dodge2022measuring}.
On the positive side, the model can then be used for a variety of applications without further debiasing, assuming that debiasing transfers~\citep{jin-etal-2021-transferability}. However, this assumption is %
under scrutiny~\citep{steed-etal-2022-upstream}. 

Intermediate training requires less energy\footnote{\citet{dodge2022measuring} report 3.1 kWh to fine-tune BERT small on MNLI, 10x less energy than pre-training}~(\fire\fire) as %
PLMs have already acquired %
representation capabilities. However, typically, all parameters are adjusted (e.g., 110M for BERT), and the question of
transferability still applies. 

Debiasing in the fine-tuning stage seems the most energy efficient (\fire). Still, all parameters must be adjusted and the additional objective and data preparation lead to increased costs. The obvious disadvantage is that for each downstream task and stereotype, debiasing needs to be conducted. \citet{lauscher2021sustainable} propose debiasing adapters. They require less energy in the debiasing procedure (\fire), but add a small overhead at inference time (ca. 1\% more parameters). Whether or not they add overhead to the fine-tuning depends on whether developers tune the whole architecture.%

Overall, we encourage NLP practitioners to consider the energy efficiency of their debiasing approach in addition to the effectiveness and usability. Energy and emission estimation tools can be used to better estimate the environmental impact of proposed approaches~\citep[e.g.,][]{lacoste2019quantifying}.

\section{How Sustainability Can Harm Fairness}
\newcite{xu2022can} hint at the potential of model compression to improve fairness. This finding holds promise for bridging the two fields. Unfortunately, the authors partially use pre-distilled models (for which they cannot control the experimental setup), do not systematically investigate the important dimensions of compression (e.g., hidden size and initialization), and do not address the stochasticity of the training procedure. In contrast, \al{\citet{silva-etal-2021-towards} and \citet{ahn-etal-2022-knowledge} demonstrate distilled models to be more biased, but either use off-the-shelf models, too, or focus on single bias dimensions and measures only.} \citet{gupta-etal-2022-mitigating} start from the assumption that compression results in unfair models and show it for one setup. We provide the first thorough analysis of compression (using the example of knowledge distillation~\citep[KD;][]{hinton2015distilling}, employing multiple tasks, bias dimensions, and measures) and show that some of these previous assumptions do not hold.

\subsection{Knowledge Distillation}
The underlying idea of knowledge distillation~\citep[KD;][]{bucilua2006compression, hinton2015distilling} is to transfer knowledge from a (typically big, pre-trained, and highly regularized) \emph{teacher} model to a (typically much smaller and untrained) \emph{student} network. It has been shown that a student network which can learn from the teacher's knowledge is likely to perform better than a small model trained without a teacher's guidance. The knowledge transfer happens through effective supervision from the teacher, e.g., via comparing output probabilities~\citep[e.g.,][]{hinton2015distilling}, comparing the intermediate features~\citep[e.g.,][]{ji2021show}, and initializing the student's layers from the teacher's layers.

\subsection{Experimental Setup}
Throughout, we use the following setup.

\paragraph{Distillation Tasks, Data Sets, and Measures.} 
We test the effects of KD on two distillation tasks: 
1)~natural language inference (NLI) using the MNLI data set~\citep{williams-etal-2018-broad}, and 
2)~semantic textual similarity (STS) prediction with the Semantic Textual Similarity-Benchmark~\citep[STS-B;][]{cer-etal-2017-semeval} data set. We chose these tasks since they are popular examples of downstream natural language understanding (NLU) tasks. There are also dedicated bias evaluation data sets and measures for the resulting models. For MNLI, we report the accuracy, and for STS the combined correlation score (average of the Pearson's correlation coefficient and Spearman's correlation coefficient).

\paragraph{Fairness Evaluation.} \al{Given that some of the existing measures have been shown to be brittle~\citep[e.g.,][]{ethayarajh-etal-2019-understanding}}, we ensure the validity of our results by combining \emph{intrinsic} with \emph{extrinsic} measures for assessing stereotypical biases along four dimensions (\emph{gender}, \emph{race}, \emph{age}, and \emph{illness}). %

\vspace{0.5em}
\noindent \emph{Word Embedding Association Test~\citep[WEAT;][]{Caliskan_2017}}.  WEAT is an intrinsic bias test that computes the differential association between two sets of target terms $A$ (e.g., \emph{woman}, \emph{girl}, etc.), and  $B$ (e.g., \emph{man}, \emph{boy}, etc.), and two sets of stereotypical attribute terms $X$ (e.g., \emph{art}, \emph{poetry}, etc.), and $Y$  (e.g., \emph{science}, \emph{math}, etc.) based on the mean similarity of their embeddings:
\vspace{-0.7em}

{\footnotesize
\begin{equation}
        w(A,B,X,Y) = \sum_{a \in A}{s(a, X, Y)} - \sum_{b \in B}{s(b, X, Y)}\,,
\end{equation}}

\normalsize
\noindent with the association $s$ of term $t\in A$ or $t\in B$ as 
\vspace{-0.5em}
{\footnotesize
\begin{equation}
    s(t,\hspace{-0.2em}X\hspace{-0.2em},\hspace{-0.2em}Y)\hspace{-0.2em}=\hspace{-0.2em} \frac{1}{|X|}\hspace{-0.5em}\sum_{x \in X}{\cos(\mathbf{t}, \mathbf{x})}  -  \frac{1}{|Y|}\hspace{-0.5em}\sum_{y \in Y}{\cos(\mathbf{t}, \mathbf{y})} \,.
\end{equation}}%
\vspace{-0.5em}

\normalsize
\noindent The final score is the effect size, computed as
\vspace{-0.5em}

{\footnotesize
\begin{equation}
\frac{\mu\hspace{-0.1em}\left(\{s(a, X, Y)\}_{a \in A}\right) - \mu\hspace{-0.1em}\left(\{s(b, X, Y)\}_{b \in B}\right)}{\sigma\left(\{s(t, X, Y)\}_{t \in A \cup B}\right)}\,,
\end{equation}}%
\vspace{-0.5em}

\normalsize
\noindent where $\mu$ is the mean and $\sigma$ is the standard deviation. To apply the measure, we follow \citet{lauscher2021sustainable}, and extract word embeddings from the PLM's encoder, using the procedure proposed by \citet{vulic-etal-2020-multi}. We use WEAT tests 3--10\footnote{WEAT tests 1 and 2 consist of bias types which do not consider marginalized social groups (flowers vs. insects, and weapons vs. music instruments)} which reflect racial (tests 3--5), gender (tests 6--8), illness (test 9), and age bias (test 10).

\vspace{0.5em}
\noindent \emph{Sentence Embedding Association Test~\citep[SEAT;][]{may-etal-2019-measuring}}. SEAT measures stereotypical bias in sentence encoders following the WEAT principle. However, instead of feeding words into the encoder, SEAT contextualizes the words of the test vocabularies via simple neutral sentence templates, e.g., \emph{``This is <word>.''}, \emph{``<word> is here.''}, etc. Accordingly, the final score is then based on comparing sentence representations instead of word representations. 
We use SEAT with the WEAT test vocabularies from tests 3--10, as before. Additionally, we use SEAT's additional Heilman Double Bind~\citep{heilman2004penalties} Competent and Likable tests which reflect gender bias, and SEAT's Angry Black Woman Stereotype~\citep[e.g.,][]{doi:10.1080/14791420903063810} test, which reflects racial bias.

\vspace{0.5em}
\noindent \emph{Bias-STS~\citep{webster2020measuring}}. The first extrinsic test is based on the Semantic Textual Similary-Benchmark~\citep[STS-B;][]{cer-etal-2017-semeval}. The idea is to measure whether a model assigns a higher similarity to a stereotypical sentence pair $s_s=(s_{s1}, s_{s2})$  than to a counter-stereotypical pair $s_c=(s_{c1}, s_{c2})$. \citet{webster2020measuring} provide templates (e.g., \emph{``A [fill] is walking.''}), which they fill with opposing gender identity terms (e.g., \emph{man}, \emph{woman}) and a profession term (e.g., \emph{nurse}) from \citet{rudinger-etal-2018-gender}  to obtain 16,980 gender bias test instances consisting of two sentence pairs (e.g., \emph{``A man is walking''} vs. \emph{``A nurse is walking''} and \emph{``A woman is walking''} vs. \emph{``A nurse is walking''}). We train the models on the STS-B training portion and collect the predictions on the created Bias-STS test set. We then follow \citet{lauscher2021sustainable} and report the \emph{average  absolute difference} between the similarity scores of male and female sentence pairs.

\vspace{0.5em}
\noindent \emph{Bias-NLI~\citep{dev2020measuring}.} Bias-NLI is another template-based test set, which allows for measuring the tendency of models to produce unfair stereotypical inferences in NLI. We train models on the MNLI training portions, and collect the predictions on the data set. It contains $1,936,512$ instances, which we create using the authors' original code as follows: we start from templates (\emph{``The <subject> <verb> a/an <object>}'') and fill the the verb and object slots with activities (e.g., \emph{``bought a car''}). To obtain a premise we fill the subject slot with an occupation (e.g., \emph{``physician''}), and to obtain the hypothesis, we provide a gendered term as the subject (e.g., \emph{``woman''}). The obtained premise-hypothesis pair (e.g., ``\textit{physician bought a car}'', ``\textit{woman bought a car}'') is \emph{neutral}, as we can not make any assumption about the gender of the premise-subject. Accordingly, we can measure the bias in the model with the \emph{fraction neutral} (FN) score --- the fraction of examples for which the model predicts the \emph{neutral} class --- and as \textit{net neutral} (NN) --- the average probability that the model assigns to the \emph{neutral} class across all instances. Thus, in contrast to the other measures, a higher FN or NN value indicates lower bias.

\paragraph{Models and Distillation Procedure.} We start from BERT~\citep{devlin-etal-2019-bert} in \emph{base} configuration (12 hidden layers, 12 attention heads per layer, hidden size of 768) available on the Huggingface hub~\citep{wolf-etal-2020-transformers}.\footnote{\url{https://huggingface.com}} We obtain teacher models from the PLM by optimizing BERT's parameters on the training portions of the respective data sets. We train the models with Adam~\citep{AdamW} (cross-entropy loss for MNLI, mean-squared error loss for STS-B) for maximum $10$ epochs and apply early stopping based on the validation set performance (accuracy for MNLI, combined correlation score for STS-B) with a patience of $2$ epochs. We grid search for the optimal batch size $b_t \in \{16, 32\}$ and learning rate $\lambda_t \in \{2 \cdot 10^{-5}, 3 \cdot 10^{-5}, 5 \cdot 10^{-5}\}$. For ensuring validity of our results~\citep{ReportingScore} we conduct this procedure 3 times starting from different random initializations. As a result, for each of the two tasks, we obtain 3 optimized teacher models. %
For all distillation procedures, we use the TextBrewer~\citep{yang-etal-2020-textbrewer} framework's \emph{GeneralDistiller}. We optimize the following hyperparameters:  batch size $b_d \in \{64, 128\}$ and temperature $t_d \in \{4, 8\}$. We distill for maximum 60 epochs and apply early stopping based on the validation score with a patience of 4 epochs. 
If we initialize the students' layers, we only apply the task-specific loss on the difference between the teacher's and the student's output. If no layers are initialized, we add a layer matching loss based on Maximum Mean Discrepancy~\citep{huang2017like}.
We use Adam with a
learning rate of $1\cdot 10^{-4}$ (warm up over 10\% of the total number of steps and linearly decreasing learning rate schedule).

\begin{figure*}[th!]
     \centering
     \begin{subfigure}[b]{0.329\textwidth}
         \centering
         \includegraphics[width=\textwidth, trim=0.6em 0.3em 3em 0.3em, clip]{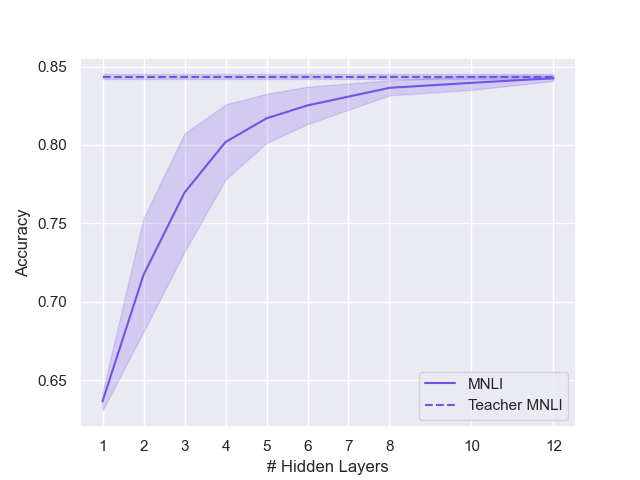}
         \caption{MNLI}
         \label{fig:mnli_layers_acc}
     \end{subfigure}
     \begin{subfigure}[b]{0.329\textwidth}
         \centering
         \includegraphics[width=\textwidth, trim=0.6em 0.3em 3em 0.3em, clip]{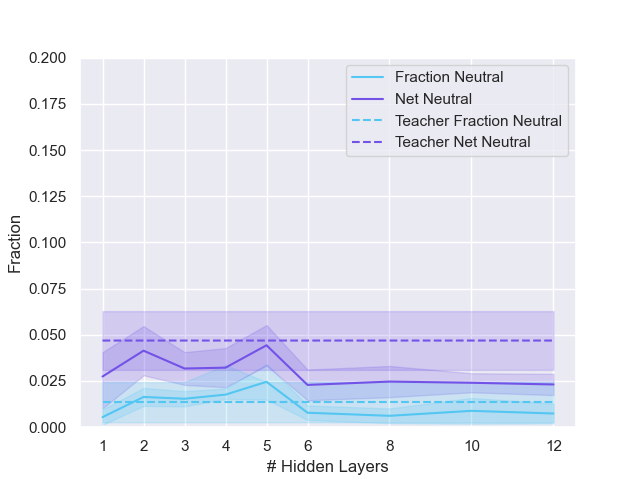}
         \caption{Bias-NLI (Gender)}
         \label{fig:mnli_layers_biasnli}
     \end{subfigure}
          \begin{subfigure}[b]{0.329\textwidth}
         \centering
         \includegraphics[width=\textwidth, trim=0.6em 0.3em 3em 0.3em, clip]{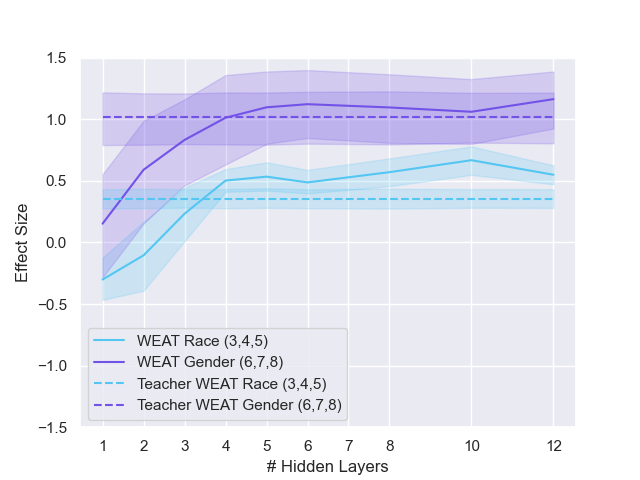}
         \caption{WEAT (Gender \& Race)}
         \label{fig:mnli_layers_weatgr}
     \end{subfigure}
    \begin{subfigure}[b]{0.329\textwidth}
         \centering
         \includegraphics[width=\textwidth, trim=0.6em 0.3em 3em 0.3em, clip]{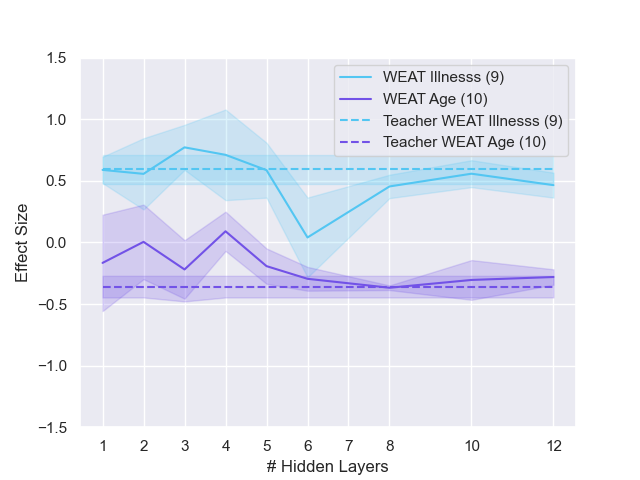}
         \caption{WEAT (Age \& Illness)}
         \label{fig:mnli_layers_weatai}
     \end{subfigure}
          \begin{subfigure}[b]{0.329\textwidth}
         \centering
         \includegraphics[width=\textwidth, trim=0.6em 0.3em 3em 0.3em, clip]{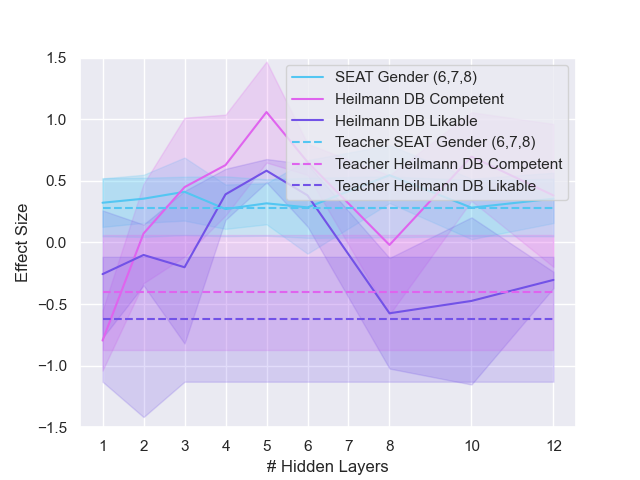}
         \caption{SEAT (Gender)}
         \label{fig:mnli_layers_seatg}
     \end{subfigure}
               \begin{subfigure}[b]{0.329\textwidth}
         \centering
         \includegraphics[width=\textwidth, trim=0.6em 0.3em 3em 0.3em, clip]{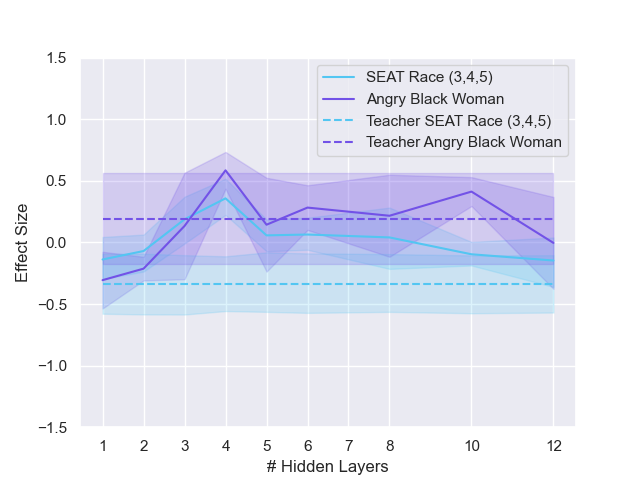}
         \caption{SEAT (Race)}
         \label{fig:mnli_layers_seatr}
     \end{subfigure}
    \caption{Results for our KD analysis (number of student hidden layers) on MNLI without initialization of the layers. We depict (a) the accuracy on MNLI, (b) the fraction neutral and net neutral scores on Bias-NLI, (c) WEAT effect sizes averaged over tests 3--5 (race) and 6--8 (gender), \al{(d) WEAT effect sizes for tests 9 (illness) and 10 (age), (e) SEAT effect sizes averaged over tests 6--8 (gender) and the Heilmann Double Bind tests, and (f) SEAT effect sizes for tests 3--5 (race) and the Angry Black Woman stereotype test}. All results are shown as average with 90\% confidence interval for the 3 teacher models (dashed lines) and 1--12 layer student models distilled from the teachers.}
    \label{fig:mnli_layers}
\end{figure*}
\paragraph{Dimensions of Analysis.} We focus on 3 dimensions: (1) we test the effect of reducing the \emph{number of layers} of the student model and report results on students with 12--1 hidden layers for MNLI and 10--1 hidden layers for STS. All other parameters stay fixed: we set the hidden size to 768 and the number of attention heads per layer to 12 (as in the teacher). We either initialize all layers of the student randomly (for MNLI)\footnote{Not mapping the layers, i.e., random initialization, yielded sub par performance for STS} or map teacher's layers to student layers for the initialization (for MNLI and STS) according to the scheme provided in the Appendix. (2) The number of layers corresponds to a \emph{vertical} reduction of the model size. Analogously, we study \emph{horizontal} compression  reflected by the \emph{hidden size} of the layers. We analyze bias in students with a hidden size $h \in [768, 576, 384, 192, 96]$. Here, we fix the number of hidden layers to 4. We follow \citet{turc2019well} and set the number of self-attention heads to $h/64$ and the feed-forward filter-size to $4h$. (3) Finally, we test the effect of the \emph{layer initialization}. To this end, we constrain the student model to have 4 hidden layers, and a hidden size of 768. We then initialize each of the students layers $l_s \in [0,4]$ (where 0 is the embedding layer) either \emph{individually} or all together with the teacher's layers $l_t \in [0,12]$ for each experiment with the following mapping ($l_t \rightarrow l_s$): $0 \rightarrow 0, 3 \rightarrow 1$, $6 \rightarrow 2$, $9 \rightarrow 3$, and $12 \rightarrow 4$.   For all dimensions, we compare the students' scores with the ones of the teacher model.

\begin{figure*}[t!]
     \centering
     \begin{subfigure}[b]{0.329\textwidth}
         \centering
         \includegraphics[width=\textwidth, trim=0.6em 0.3em 3em 0.3em, clip]{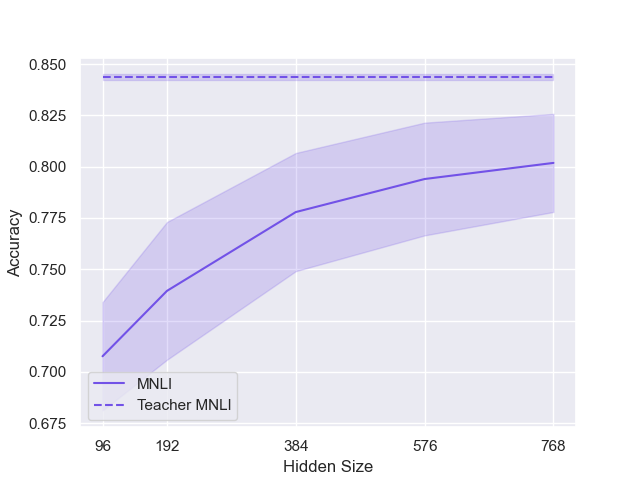}
         \caption{MNLI}
         \label{fig:mnli_hs_accuracy}
     \end{subfigure}
     \begin{subfigure}[b]{0.329\textwidth}
         \centering
         \includegraphics[width=\textwidth, trim=0.6em 0.3em 3em 0.3em, clip]{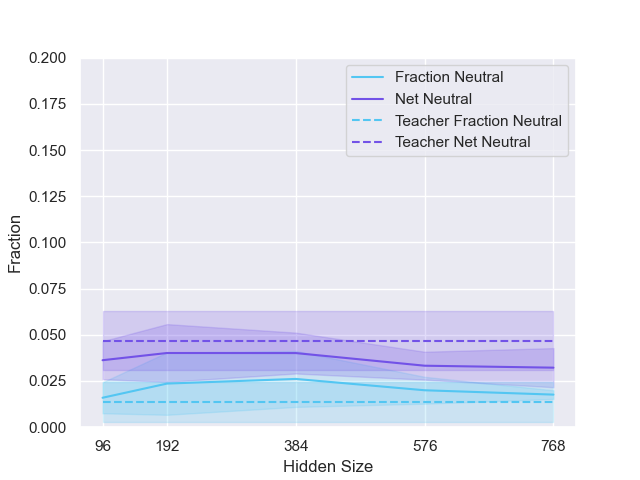}
         \caption{Bias-NLI (Gender)}
         \label{fig:mnli_hs_biasnli}
     \end{subfigure}
          \begin{subfigure}[b]{0.329\textwidth}
         \centering
         \includegraphics[width=\textwidth, trim=0.6em 0.3em 3em 0.3em, clip]{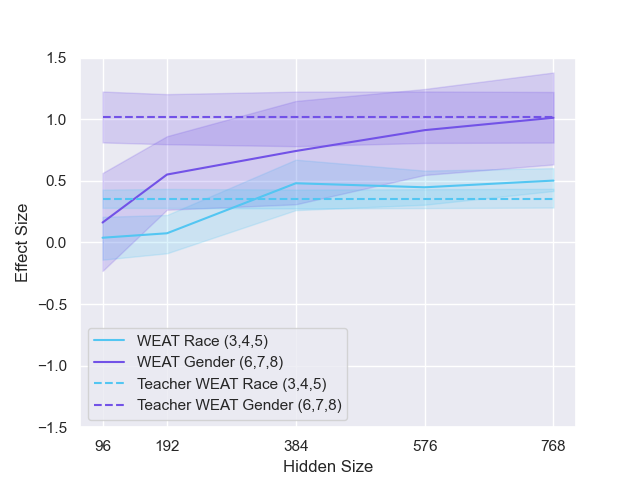}
         \caption{WEAT (Gender \& Racism)}
         \label{fig:mnli_hs_weatgr}
     \end{subfigure}
     
     \begin{subfigure}[b]{0.329\textwidth}
         \centering
         \includegraphics[width=\textwidth, trim=0.6em 0.3em 3em 0.3em, clip]{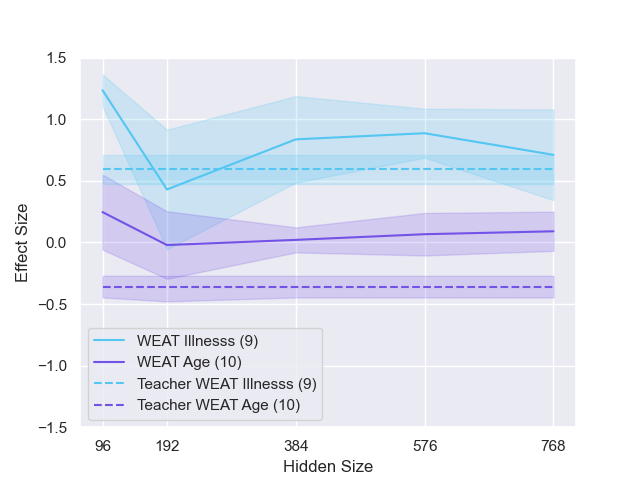}
         \caption{WEAT (Age \& Illness)}
         \label{fig:mnli_hs_weatai}
     \end{subfigure}
          \begin{subfigure}[b]{0.329\textwidth}
         \centering
         \includegraphics[width=\textwidth, trim=0.6em 0.3em 3em 0.3em, clip]{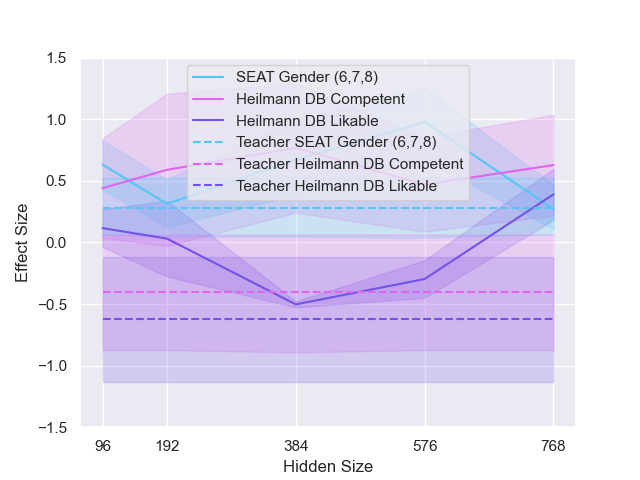}
         \caption{SEAT (Gender)}
         \label{fig:mnli_hs_seatg}
     \end{subfigure}
               \begin{subfigure}[b]{0.329\textwidth}
         \centering
         \includegraphics[width=\textwidth, trim=0.6em 0.3em 3em 0.3em, clip]{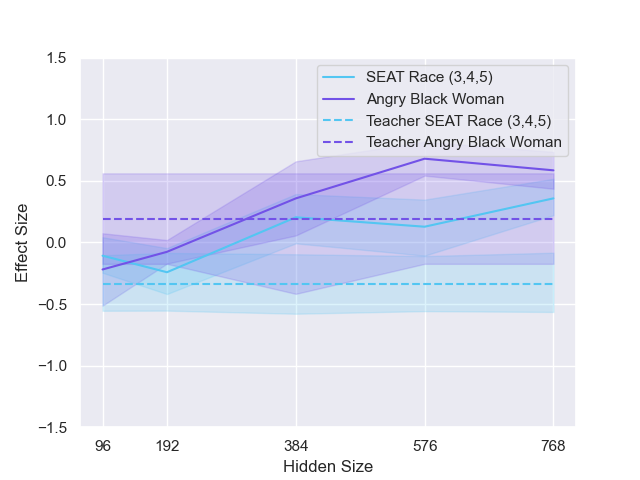}
        \caption{SEAT (Race)}
         \label{fig:mnli_hs_seatr}
     \end{subfigure}
    \caption{Results for our KD analysis (varying hidden size) on MNLI (without initialization of student layers, 4 hidden layers). We depict (a) the  accuracy on MNLI, (b) the fraction neutral and net neutral scores on Bias-NLI, (c) WEAT effect sizes averaged over tests 3,4,5 (race) and 6,7,8 (gender), \al{(d) WEAT effect sizes for tests 9 (illness) and 10 (age), (e) SEAT effect sizes averaged over tests 6--8 (gender) and the Heilmann Double Bind tests, and (f) SEAT effect sizes for tests 3--5 (race) and the Angry Black Woman stereotype test}. All results shown as average with 90\% confidence interval for the 3 teacher models (dashed lines) and 96--768 hidden size student models.}
    \label{fig:mnl_hs}
\end{figure*}
\begin{figure*}[th!]
     \centering
     \begin{subfigure}[b]{0.329\textwidth}
         \centering
         \includegraphics[width=\textwidth, trim=0.6em 0.3em 3em 0.3em, clip]{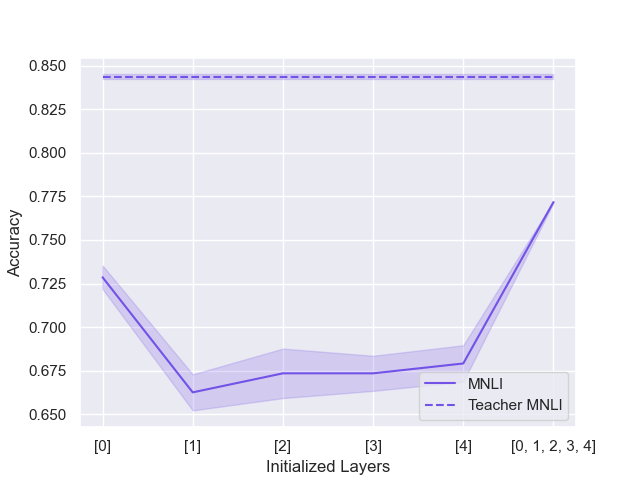}
         \caption{MNLI}
         \label{fig:mnli_init_acc}
     \end{subfigure}
     \begin{subfigure}[b]{0.329\textwidth}
         \centering
         \includegraphics[width=\textwidth, trim=0.6em 0.3em 3em 0.3em, clip]{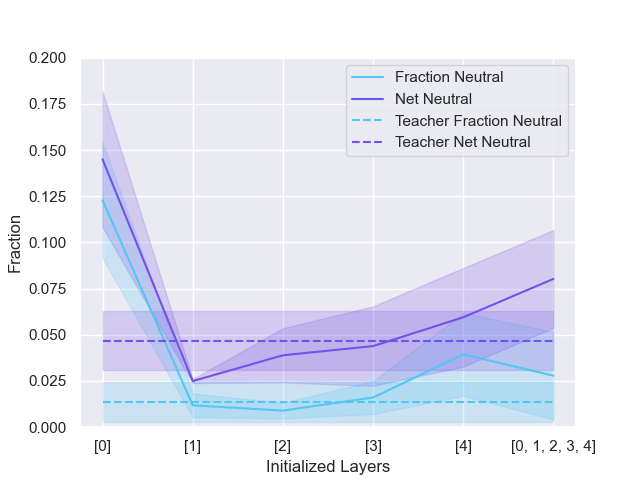}
         \caption{Bias-NLI (Gender)}
         \label{fig:mnli_init_biasnli}
     \end{subfigure}
          \begin{subfigure}[b]{0.329\textwidth}
         \centering
         \includegraphics[width=\textwidth, trim=0.6em 0.3em 3em 0.3em, clip]{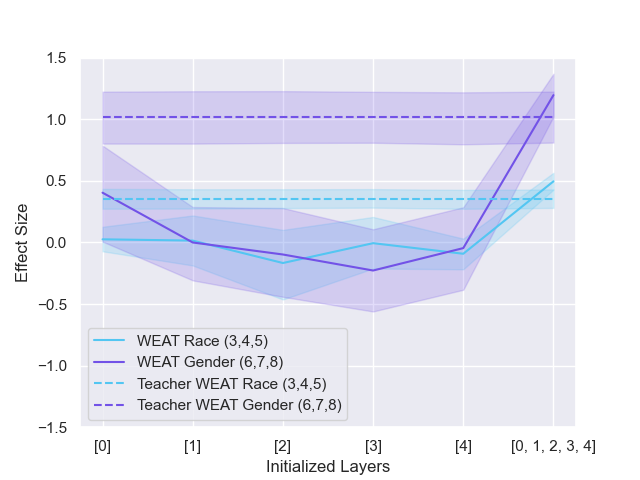}
         \caption{WEAT (Gender \& Racism)}
         \label{fig:mnli_init_weatgr}
     \end{subfigure}
    \caption{Results MNLI-KD when varying initialization of the student layers. We depict (a) the accuracy on MNLI, (b) the FN and NN scores on Bias-NLI, and (c) WEAT effect sizes averaged over tests 3,4,5 (race) and 6,7,8 (gender). All results are averages with 90\% confidence interval for the 3 teacher models (dashed lines) and students distilled from the teachers where either a single layer was initialized ([0], [1], [2], [3], or [4]) or all layers ([0, 1, 2, 3, 4]).}
    \label{fig:stilt}
\end{figure*}
\subsection{Results}
\al{We discuss the results of our KD analysis.}
\paragraph{Varying the Number of Hidden Layers.} 
Figures~\ref{fig:mnli_layers_acc}--\ref{fig:mnli_layers_seatr} show the MNLI distillation experiments, where we vary the number of student layers (without initializing them). We report the overall performance reflected by MNLI (accuracy) and the bias measured with Bias-NLI, WEAT (Tests 3--10), and SEAT (Tests 3--8, Heilman Double Bind Competent and Likable, and Angry Black Woman Stereotype). We provide the additional SEAT results (Tests 9 and 10) as well as the scores for the other tasks, STS and MNLI with initialization in the Appendix. 

The accuracy indicates that we successfully ran the distillation (Figure~\ref{fig:mnli_layers_acc}). Students with 12 hidden layers (no compression) reach roughly the same performance as their teachers. Generally, we observe that the performance variation among students is higher than among teachers, with the highest variation for students with 3 to 5 hidden layers. 

Looking at the bias measures (see Figures~\ref{fig:mnli_layers_biasnli}--\ref{fig:mnli_layers_seatr}), we note that the variation of the scores is even higher, especially among the teacher models. This observation suggests lower numerical stability of the bias measures tested. (The test set for Bias-NLI contains $\sim$2 Million instances, so this aspect cannot be attributed to lower test set sizes). Unsurprisingly, the bias results of the students are generally in roughly the same areas than the ones of their teachers. \textit{This shows that students inherit their teachers biases in the distillation process}. Grouping the test results by measure (e.g., WEAT, etc.) and dimension (e.g., race) results in roughly the same patterns of biases measurable. E.g., in Figure~\ref{fig:mnli_layers_seatr}, the results of the aggregated tests 3, 4, and 5 follow the same pattern as the Angry Black Woman Stereotype test. We hypothesize that this is due to the partially overlapping term sets. However, across measures and dimensions we find roughly the same bias behavior: students with 12 to 6 hidden layers often exhibit a higher bias than their teachers (for NLI, this corresponds to a lower FN)! The exception to this rule is WEAT test 9, illness. For most tests, the highest bias arises with 4 hidden layers. Students with lower number of layers are mostly less biased across all tests. However, from this point on, the accuracy on MNLI also drops more strongly. These findings are in stark contrast to the results of \newcite{xu2022can}.

\paragraph{Varying the Hidden Size.} We show the results of KD when varying the hidden size of the student models (number of hidden layers is fixed to 4) in Figures~\ref{fig:mnli_hs_accuracy}--\ref{fig:mnli_hs_seatr}. As before, we provide additional results in the Appendix. Generally, we note that the performance curve (Figure~\ref{fig:mnli_hs_accuracy}) is again in-line with our expectations. %
As in the previous experiment we note high variations of the scores and students biases mostly seem to be located in roughly the same ball park as their teachers' scores. However, we again note that the concrete behavior of the bias curves depends on bias measure and dimension. Interestingly, the curves when varying the hidden size look (with some exceptions) similar to the ones when varying the number of hidden layers. We thus hypothesize, that \emph{both vertical and horizontal compression have a similar affect on fairness}.

\paragraph{Varying the Initialization.} As a last aspect of our analysis, we look at the effect of initializing various layers of the student with the weights of the teacher. We depict some of the scores in Figures~\ref{fig:mnli_init_acc}--\ref{fig:mnli_init_weatgr}. Interestingly, changing the initialization has a large effect both on the MNLI accuracy, as well as on the bias measures. These findings highlight again that \emph{monitoring fairness after KD is crucial}.

\vspace{0.5em}
Overall, our findings show that the devil is in the detail. While generally, \textbf{the amount of bias in the distilled models is inherited from the teacher's biases} and \textbf{the biases measurable seem to roughly group by social bias dimension and measure}, biases still need to be carefully tested. Most importantly, while \citet{xu2022can} point at the potential of KD for increasing fairness, we cannot confirm this observation. In contrast, \textbf{across most bias measures tested, the student models start from a higher amount of bias than the teacher}. \al{A possible explanation for this behavior is that \emph{weak learners}, i.e., models with limited capacity, generally show a stronger tendency to exploit biases in the data set during the learning process than models with higher capacity~\citep{sanh2020learning}.}

\section{Related Work}

\vspace{0.3em}
\noindent\textbf{Fairness in NLP.} There exists a plethora of works on increasing the fairness of NLP models, most prominently focused on the issue of unfair stereotypes in the models~\citep[e.g.,][\emph{inter alia}]{Caliskan_2017,zhao-etal-2017-men,dev2020measuring, nadeem2020stereoset}. We only provide an overview and refer the reader to more comprehensive surveys on the topic~\citep[e.g.,][]{sun-etal-2019-mitigating,blodgett2020, shah-etal-2020-predictive}.
\citet{Bolukbasi} were the first to point to the issue of stereotypes encoded in static word embeddings, which led to a series of works focused on measuring and mitigating these biases~\citep[e.g.,][]{dev2019attenuating, DEBIE}, as well as assessing the reliability of the tests~\citep{gonen-goldberg-2019-lipstick,ethayarajh-etal-2019-understanding,antoniak-mimno-2021-bad,delobelle2021measuring, blodgett-etal-2021-stereotyping}. For instance, \citet[][]{Caliskan_2017} proposed the well-known WEAT. Recent works focus on measuring and mitigating bias in contextualized language representations~\citep{kurita-etal-2019-measuring, bordia-bowman-2019-identifying,qian-etal-2019-reducing, webster2020measuring, nangia-etal-2020-crows, sap-etal-2020-social} and in downstream scenarios, e.g., for dialog~\citep[e.g.,][]{thewomanworked,dinan-etal-2020-queens, barikeri-etal-2021-redditbias}, co-reference resolution~\citep{zhao-etal-2018-gender}, and NLI~\citep{rudinger-etal-2017-social,dev2020measuring}. Similarly, researchers have explored multilingual scenarios~\citep[e.g.,][]{lauscher-glavas-2019-consistently, lauscher-etal-2020-araweat, ahn-oh-2021-mitigating}, more fine-grained biases~\citep{dinan-etal-2020-multi}, and more biases, beyond the prominent sexism and racism dimensions~\citep[e.g.,][]{zhao-etal-2018-gender,rudinger-etal-2018-gender}, like speciesist bias~\citep{takeshita2022speciesist}.

\vspace{0.3em}
\noindent\textbf{Sustainability in NLP.} \citet{strubell-etal-2019-energy} have called for more awareness of NLP's environmental impact. Reducing the energy consumption can be achieved through efficient pre-training~\citep{di2021efficient}, smaller models and employing less pre-training data considering the specific needs of the task at hand~\citep[e.g.,][]{perez-mayos-etal-2021-much,zhang-etal-2021-need}. If a PLM is already in-place, one can rely on sample-efficient methods~\citep[e.g.,][]{lauscher-etal-2020-zero}, or refrain from fully fine-tuning the model~\citep[e.g,][]{Adapters, pfeiffer2021adapterfusion}. Similarly, one can compress the models via distillation~\citep[e.g.,][]{hinton2015distilling, sanh2019distilbert, he-etal-2021-distiller}, pruning~\citep[e.g.,][]{fan2019reducing, li-etal-2020-efficient-transformer, wang-etal-2020-structured}, and quantization~\citep[e.g.,][]{zhang-etal-2020-ternarybert}, to increase energy-efficiency of later training stages or at inference time. A survey is provided by~\citet{10.1145/3487045}. In the area of distillation, researchers have explored distillation in different setups, e.g., for a specific task~\citep[e.g.,][]{see-etal-2016-compression}, on a meta-level~\citep[e.g.,][]{he-etal-2021-distiller}, or for a specific resource scenario~\citep[e.g.,][]{wasserblat-etal-2020-exploring}. 
Other efforts focused on accurate energy and emission measurement and provide tools for monitoring energy consumption~\citep[e.g.,][]{lacoste2019quantifying,cao-etal-2020-towards}. While most research in the area of NLP focuses on reducing operational costs, i.e.,\,carbon emissions due to the energy required to develop and run models, downstream impacts of model deployment stand to have a much larger impact on the environment~\citep{kaack:hal-03368037}. See \citet{ccai-22-tackling} for a detailed presentation of how machine learning can help to counter climate change more broadly, including a disucussion of NLP applications. 

\vspace{0.3em}
\noindent\textbf{Bridging Fairness and Sustainability.} To the best of our knowledge, there are currently only few works that are located at the intersection of the two fields in NLP: \citet{lauscher2021sustainable} proposed to use adapters for decreasing energy consumption during training-based debiasing and increasing the reusability of this knowledge, \al{which has been proven effective by \citet{holtermann-etal-2022-fair}}. Recently, the unpublished work of \citet{xu2022can} asks whether compression can improve fairness. \al{In contrast, \citet{silva-etal-2021-towards} find that off-the-shelf distilled models, such as DistilBERT, exhibit higher biases, but do not provide a systematic evaluation of the effect of KD dimensions. Concurrent to our work,  \citet{ahn-etal-2022-knowledge} demonstrate similar trends, but focus on gender bias (quantified through a single measure) and the number of hidden layers in the student, only. }
Starting from the assumption that compression can lead to biased models, \citet{gupta-etal-2022-mitigating} propose a fairness-increasing KD loss and demonstrate their baselines to be more biased. In a similar vein, \citet{xu-etal-2021-beyond} discuss the robustness of BERT compression. In computer vision, researchers have shown that compression exacerbates algorithmic bias \citep[e.g.,][]{hooker2020characterising}. E.g., \citet{liebenwein2021lost} demonstrate pruned models to be brittle to out-of-distribution points. %
\citet{ahia-etal-2021-low-resource} present the most relevant work in this space, exploring %
the \textit{low-resource double-bind}: individuals with the least access to computational resources are also likely to have scarce data resources. %
They find that model pruning can %
lead to better performance on low-resource languages by reducing undesirable memorization of rare examples. This study represents a valuable step towards better understanding the intersection of fairness and sustainability. \al{In this work, we argue that more research is needed to understand the complex relationships between the two fields.} %

\section{Conclusion}
Fairness and environmental sustainability are equally important goals in NLP. However, the vast majority of research in our community focuses exclusively on one of these aspects. We argue that bridging fairness and environmental sustainability is thus still an unresolved issue. To start bringing these fields together in a more holistic research on ethical issues in NLP, we conducted a two-step analysis: first, we provided an overview on the efficiency of \al{exemplary} %
fairness approaches. Second, we ran an empirical analysis of the fairness of KD, as a popular example of methods to enhance sustainability. 
We find that use of KD can actually decrease fairness, motivating our plea for research into joint approaches.
We hope that our work inspires such research on the interplay between the two fields \al{for fair and sustainable NLP}.

\section*{Acknowledgements}
This work is in part funded by the  European Research Council under the European Union’s Horizon 2020 research and innovation program (grant agreement No. 949944, INTEGRATOR). The work of Anne Lauscher is funded under the Excellence Strategy of the Federal Government and the Länder. At the time of writing, DH and AL were members of the Data and Marketing Insights unit of the Bocconi Institute for Data Science and Analysis. We thank the anonymous reviewers for their insightful comments.

\section*{Limitations}
Our work deals with the general relationship between environmental sustainability and fairness. As a showcase, we explore the effect of KD on stereotypical bias measures. This does not imply that KD is the only or the most egregious method, and more research into other approaches is needed. 
In this context, we resorted to established bias measures, which treat gender as a binary variable. This is due to the limitations of the established data sets, some of which allow for measuring ``classic'' sexism and do not reflect the large spectrum of possible  identities~\citep{lauscher2022welcome}. We further acknowledge that, in this research, we only worked with  examples to highlight the importance of considering both sustainability goals. We acknowledge that to truly understand the relationship between fairness and environmental sustainability, it requires more in-depth studies. We thus encourage future research to explore the interplay between the   fields 
more, for other societal biases, including but not limited to queerphobia and non-binary exclusion~\citep{dev-etal-2021-harms}, and for other sustainability and fairness approaches and aspects.

\bibliography{custom}
\bibliographystyle{acl_natbib}

\clearpage
\appendix

\label{sec:appendix}
\section{Links to Data, Models, and Code Bases}
\setlength{\tabcolsep}{6pt}
\begin{table*}[th!]
\centering
\begin{tabularx}{\textwidth}{l l X}
\toprule
{\bf Purpose} & {\bf Name} & {\bf URL} \\ \midrule
Natural Language Inference Data & MNLI & \url{https://huggingface.co/datasets/glue} \\
Semantic Similarity Prediction Data & STS-B & \url{https://huggingface.co/datasets/glue} \\
\midrule
Intrinsic Bias Test Terms & WEAT & \url{https://www.science.org/doi/10.1126/science.aal4230}\\
Intrinsic Bias Test Sentences & SEAT & \url{https://github.com/W4ngatang/sent-bias} \\
Extrinsic Bias Code for Data & Bias-NLI & \url{https://github.com/sunipa/On-Measuring-and-Mitigating-} \url{Biased-Inferences-of-} \url{Word-Embeddings} \\
Extrinsic Bias Templates & Bias-STS & \url{https://arxiv.org/pdf/2010.06032.pdf} \\
\midrule
General Code Base & Transformers & \url{https://transformer.huggingface.co} \\
Code Base for Distillation & TextBrewer & \url{https://github.com/airaria/TextBrewer} \\
\bottomrule
\end{tabularx}
\caption{Links to the datasets and code bases used in our work. }
\label{tbl:datasets}
\end{table*}
We provide the links to datasets and code bases used in this work in Table~\ref{tbl:datasets}. In all distillation experiments, we start from BERT in base configuration: \url{https://huggingface.co/bert-base-uncased}. Our code is provided \al{in the GitHub repository linked in the main body of the manuscript}.

\subsection{Details on the Initialization}
When varying the hidden layers of the student and fully initializing the layers, the layer initialization is dependent on the student's size. Following common practice, the mapping is spread across the layers of the teacher. We provide the mapping here ($l_t \rightarrow l_s$):
\begin{itemize}
    \item 10 student layers: $0\rightarrow0$, $2\rightarrow1$, $3\rightarrow2$, $4\rightarrow3$, $5\rightarrow4$, $6\rightarrow5$, $7\rightarrow6$, $8\rightarrow7$, $9\rightarrow8$, $10\rightarrow9$ $12\rightarrow10$
    \item 8 student layers: $0\rightarrow0$, $2\rightarrow1$, $4\rightarrow2$, $5\rightarrow3$, $6\rightarrow4$, $7\rightarrow5$, $8\rightarrow6$, $10\rightarrow7$, $12\rightarrow8$
        \item 6 student layers: $0\rightarrow0$, $2\rightarrow1$, $4\rightarrow2$, $6\rightarrow3$, $8\rightarrow4$, $10\rightarrow5$, $12\rightarrow6$
        \item 5 student layers: $0\rightarrow0$, $3\rightarrow1$, $5\rightarrow2$, $7\rightarrow3$, $9\rightarrow4$, $12\rightarrow5$
                \item 4 student layers: $0\rightarrow0$, $3\rightarrow1$, $6\rightarrow2$, $9\rightarrow3$, $12\rightarrow4$ 
                \item 3 student layers: $0\rightarrow0$, $4\rightarrow1$, $8\rightarrow2$, $12\rightarrow3$
                \item 2 student layers: $0\rightarrow0$, $6\rightarrow1$, $12\rightarrow2$
                \item 1 student layer: $0\rightarrow0$, $6\rightarrow1$
\end{itemize}

\section{Additional Results}
We provide the additional results for our distillation experiments.
\subsection{Varying the Number of Student Layers}

\begin{figure}[t]
     \centering
    \includegraphics[width=\linewidth]{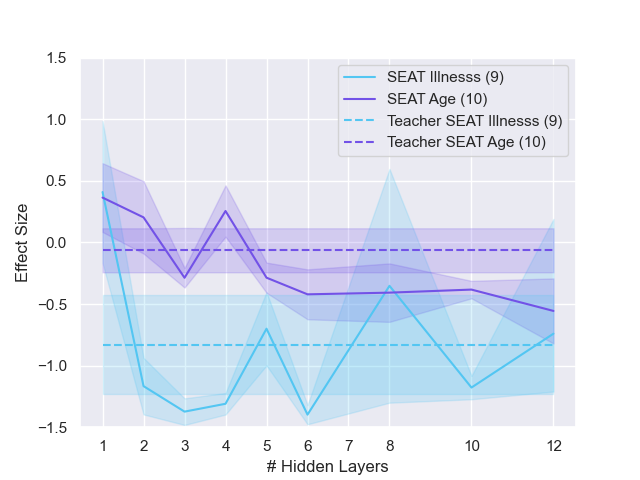}
    \caption{Additional results for our KD analysis (number of hidden layers w/o initialization). We show the MNLI distillation results for SEAT tests 9 and 10.}
    \label{fig:mnli_hl_seat_ageillness}
\end{figure}
\begin{figure}[t]
     \centering
    \includegraphics[width=\linewidth]{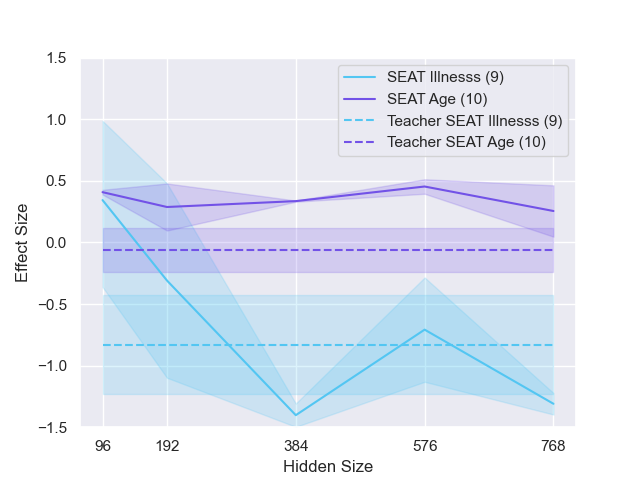}
    \caption{Additional results for our KD analysis (varying hidden size) on MNLI (without initialization of student layers, 4 hidden layers). We show the MNLI distillation results for SEAT tests 9 and 10.}
    \label{fig:mnl_hs_app}
\end{figure}

\begin{figure*}[t]
     \centering
     \begin{subfigure}[b]{0.35\textwidth}
         \centering
         \includegraphics[width=\textwidth]{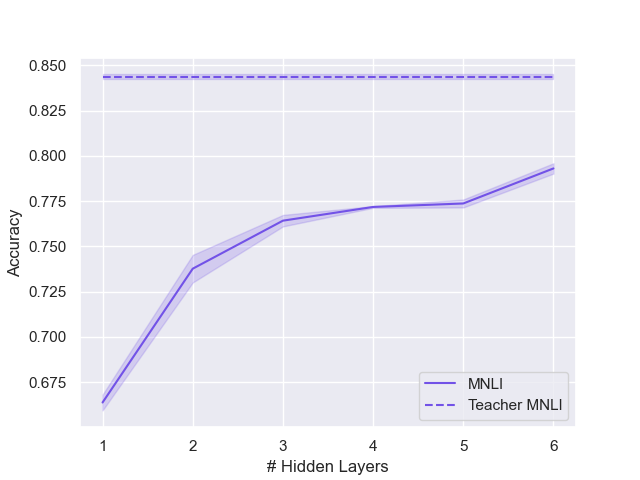}
         \caption{MNLI}
     \end{subfigure}
     \begin{subfigure}[b]{0.35\textwidth}
         \centering
         \includegraphics[width=\textwidth]{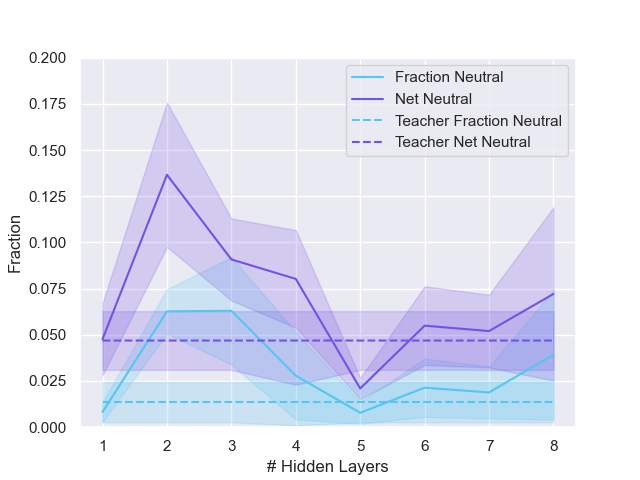}
         \caption{Bias-NLI}
     \end{subfigure}
          \begin{subfigure}[b]{0.32\textwidth}
         \centering
         \includegraphics[width=\textwidth]{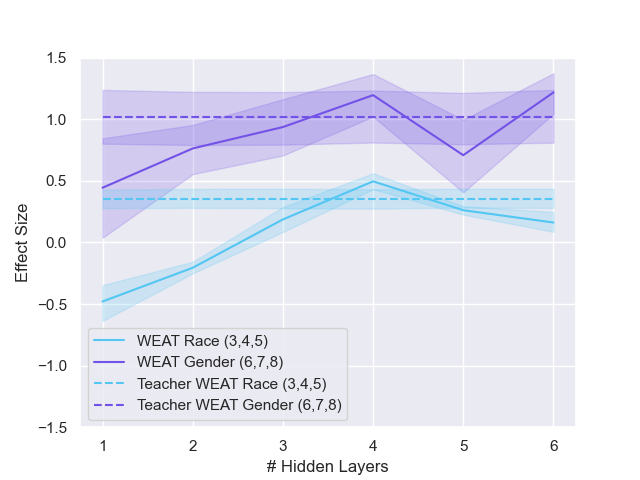}
         \caption{WEAT (Gender \& Racism)}
     \end{subfigure}
         \begin{subfigure}[b]{0.32\textwidth}
         \centering
         \includegraphics[width=\textwidth]{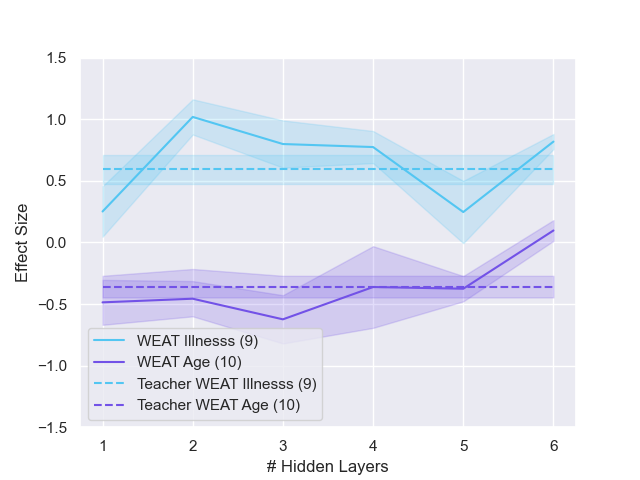}
         \caption{WEAT (Age \& Illness)}
     \end{subfigure}
          \begin{subfigure}[b]{0.32\textwidth}
         \centering
         \includegraphics[width=\textwidth]{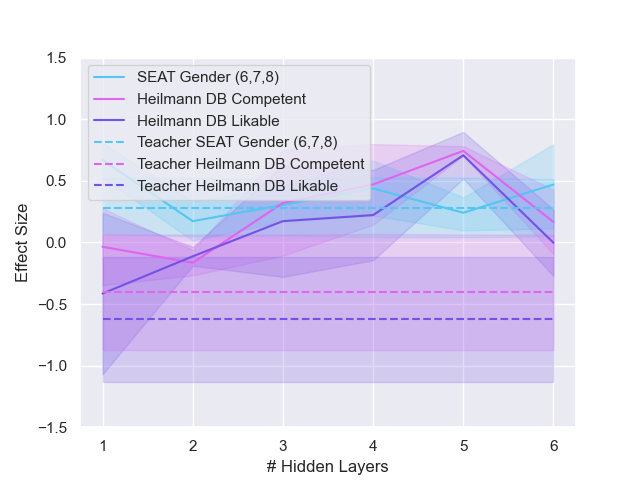}
         \caption{SEAT (Gender)}
     \end{subfigure}
               \begin{subfigure}[b]{0.32\textwidth}
         \centering
         \includegraphics[width=\textwidth]{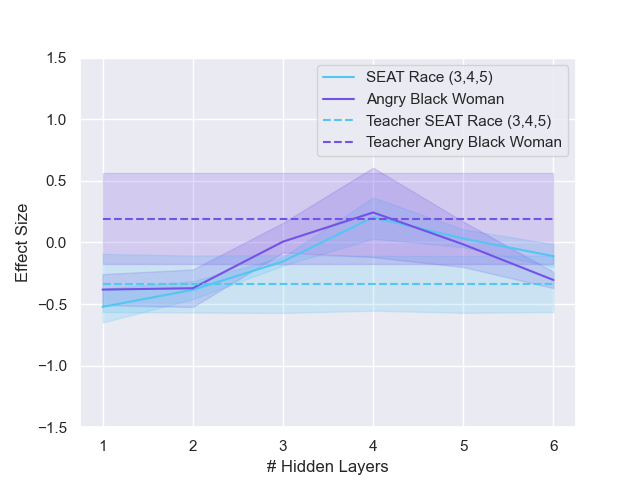}
         \caption{SEAT (Race)}
     \end{subfigure}
     \begin{subfigure}[b]{0.32\textwidth}
         \centering
         \includegraphics[width=\textwidth]{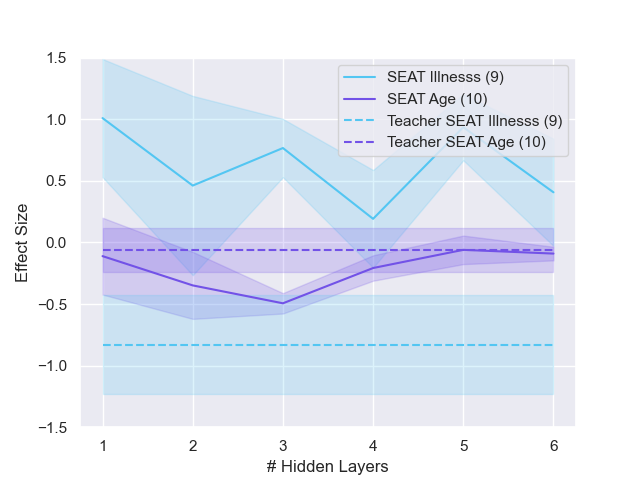}
         \caption{SEAT (Age \& Illness)}
     \end{subfigure}
    \caption{Results for our KD analysis (number of hidden layers) on MNLI distillation with initialization of all layers. We depict (a) the accuracy on MNLI, (b) the fraction neutral and net neutral scores on Bias-NLI, (c) WEAT effect sizes averaged over tests 3,4,5 (race) and 6,7,8 (gender), (d) WEAT 9 and 10 (age and illness), (e) SEAT scores for tests 6,7,8 and the Heilmann Double Bind tests (gender), (f) SEAT scores for 3,4,5 and Angry Black Woman Stereotype, and (g) SEAT 9 and 10 (age and illness). All results are shown as average with 90\% confidence interval for the 3 teacher models (dashed lines) and 1--12 layer student models distilled from the teachers.}
    \label{fig:mnli_all_layers_layers}
\end{figure*}

The additional results for the MNLI distillation, i.e., the SEAT scores for tests 9 and 10 are shown in Figure~\ref{fig:mnli_hl_seat_ageillness}. 
The MNLI results when initializing all layers are depicted in Figure~\ref{fig:mnli_all_layers_layers}.

\begin{figure*}[th!]
     \centering
         \begin{subfigure}[b]{0.49\textwidth}
         \centering
         \includegraphics[width=\textwidth]{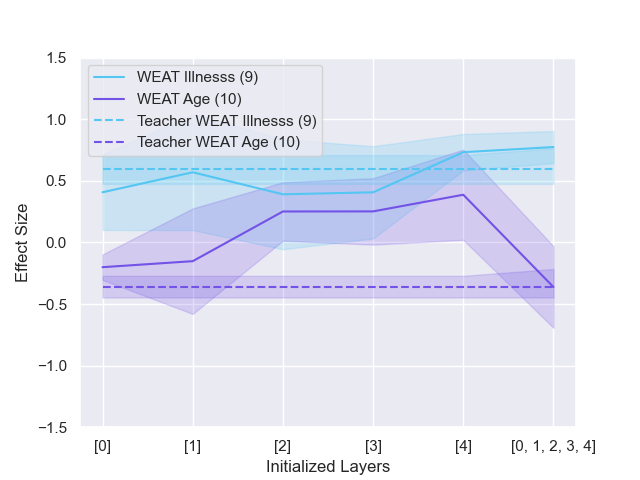}
         \caption{WEAT (Age \& Illness)}
     \end{subfigure}
          \hfill
          \begin{subfigure}[b]{0.49\textwidth}
         \centering
         \includegraphics[width=\textwidth]{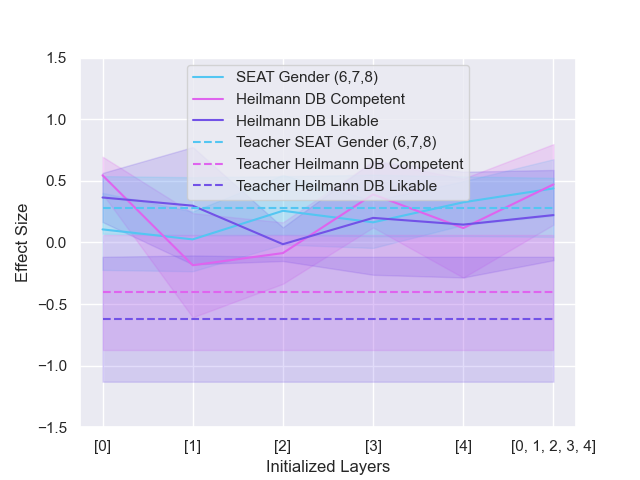}
         \caption{SEAT (Gender)}
         \label{fig:mnli_init_seat_gender}
     \end{subfigure}
               \begin{subfigure}[b]{0.49\textwidth}
         \centering
         \includegraphics[width=\textwidth]{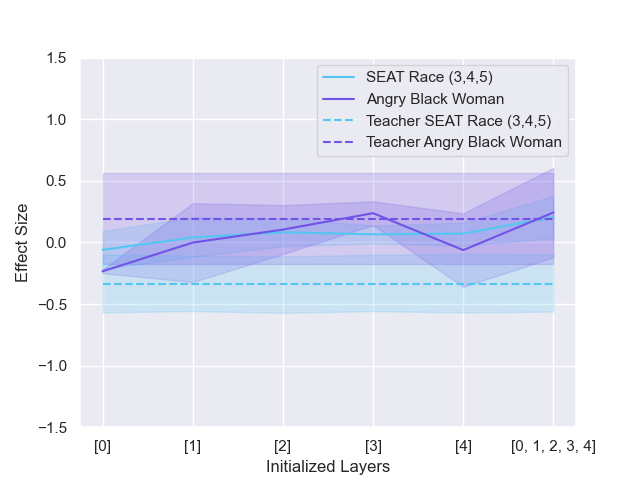}
         \caption{SEAT (Race)}

    \end{subfigure}
    \begin{subfigure}[b]{0.49\textwidth}
         \centering
         \includegraphics[width=\textwidth]{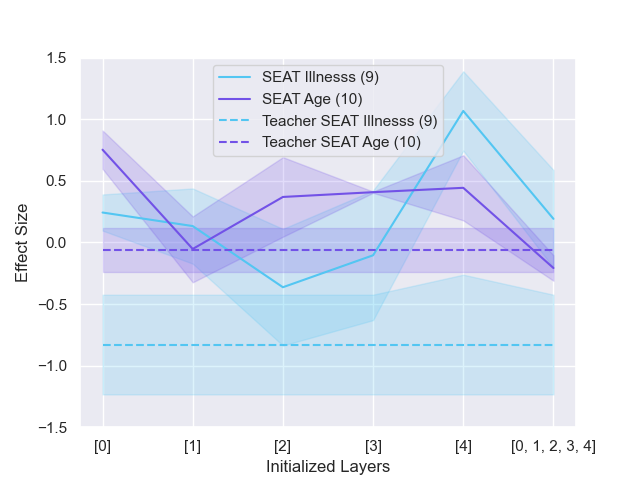}
         \caption{SEAT (Age \& Illness)}

    \end{subfigure}
    \caption{Results for our KD analysis (varying initialization of the student layers) on MNLI. We depict (a) WEAT Age \& Illness, (b) SEAT Gender, (c) SEAT Race, and (d) SEAT Age \& Illness. All results are shown as average with 90\% confidence interval for the 3 teacher models (dashed lines) and student models distilled from the teachers where either a single layer was initialized ([1], [2], [3], or [4]) or all layers ([1, 2, 3, 4]).}
    \label{fig:mnl_init}
    \vspace{-0.7em}
\end{figure*}

\begin{figure*}[t]
     \centering
     \begin{subfigure}[b]{0.35\textwidth}
         \centering
         \includegraphics[width=\textwidth]{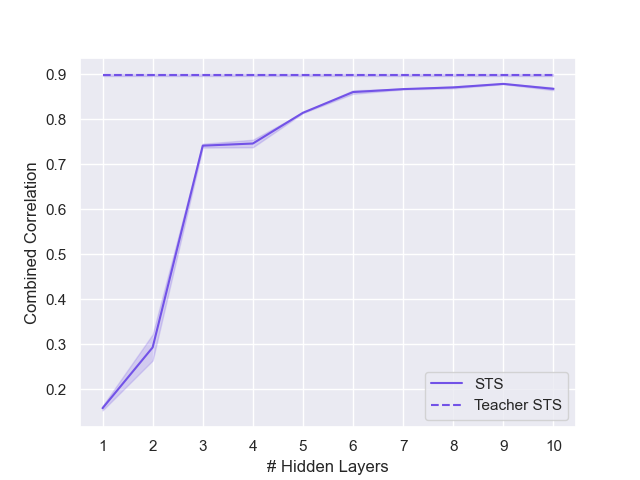}
         \caption{STS-B}
     \end{subfigure}
     \begin{subfigure}[b]{0.35\textwidth}
         \centering
         \includegraphics[width=\textwidth]{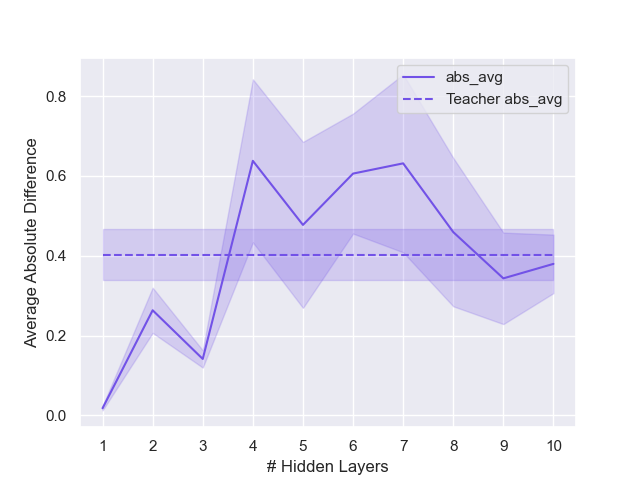}
         \caption{Bias-STS (Gender)}
     \end{subfigure}
          \begin{subfigure}[b]{0.32\textwidth}
         \centering
         \includegraphics[width=\textwidth]{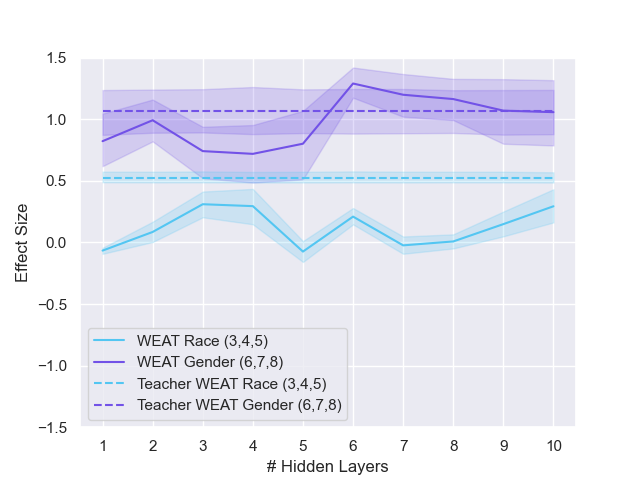}
         \caption{WEAT (Gender \& Racism)}
     \end{subfigure}
         \begin{subfigure}[b]{0.32\textwidth}
         \centering
         \includegraphics[width=\textwidth]{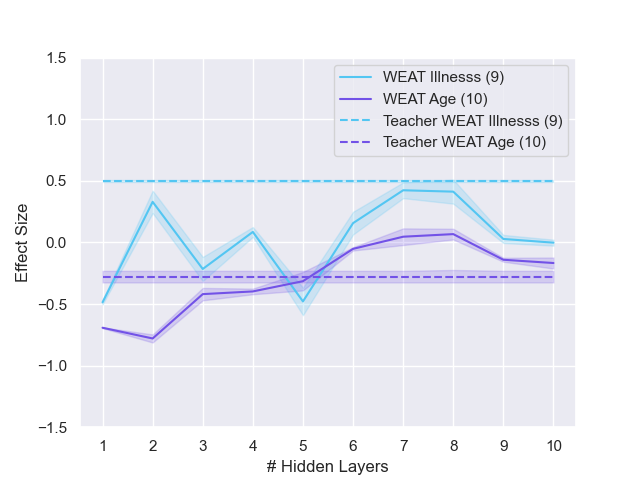}
         \caption{WEAT (Age \& Illness)}

     \end{subfigure}
          \begin{subfigure}[b]{0.32\textwidth}
         \centering
         \includegraphics[width=\textwidth]{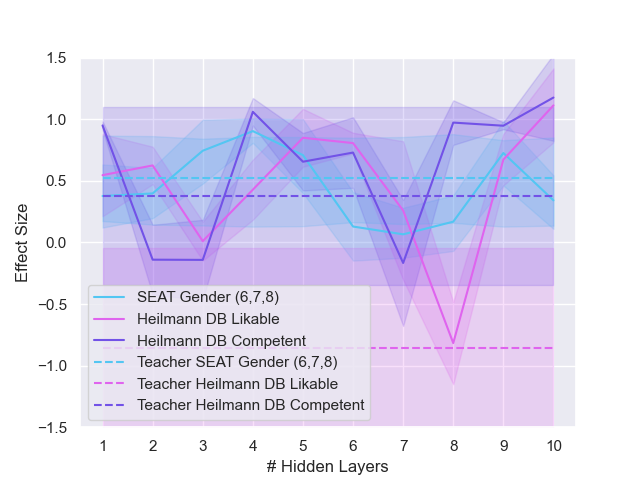}
         \caption{SEAT (Gender)}

     \end{subfigure}
               \begin{subfigure}[b]{0.32\textwidth}
         \centering
         \includegraphics[width=\textwidth]{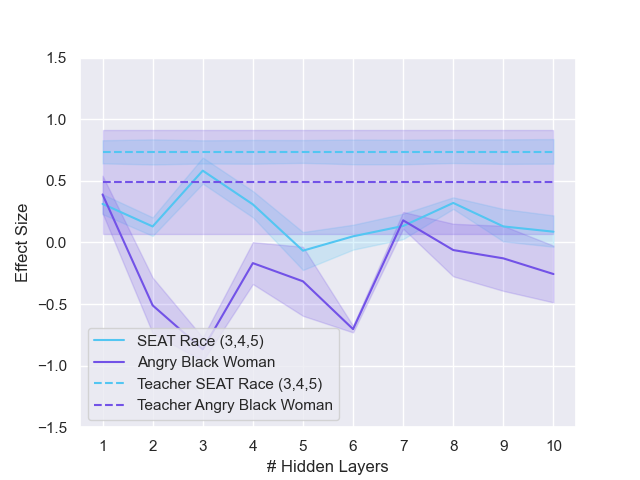}
         \caption{SEAT (Race)}
         \end{subfigure}
         \begin{subfigure}[b]{0.32\textwidth}
         \centering
         \includegraphics[width=\textwidth]{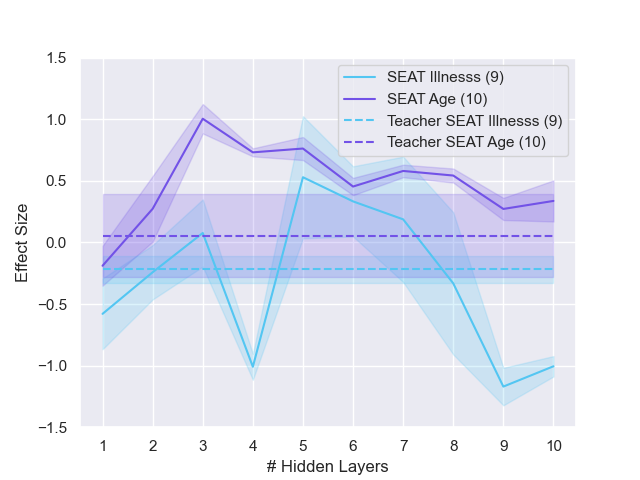}
         \caption{SEAT (Age \& Illness)}

     \end{subfigure}
    \caption{Results for our KD analysis (number of hidden layers) on STS-B distillation with initialization of all layers. We depict (a) the correlation on STS (measured as average of the Pearson and Spearman correlation coefficients), (b) the average absolute difference on Bias-STS, (c) WEAT effect sizes averaged over tests 3,4,5 (race) and 6,7,8 (gender), (d) WEAT 9 and 10 (age and illness), (e) SEAT scores for tests 6,7,8 and the Heilmann Double Bind tests (gender), (f) SEAT scores for 3,4,5 and Angry Black Woman Stereotype, and (g) SEAT 9 and 10 (age and illness)  All results are shown as average with 90\% confidence interval for the 3 teacher models (dashed lines) and 1--12 layer student models distilled from the teachers.}
    \label{fig:sts}
\end{figure*}

\subsection{Varying the Hidden Size}
We provide the additional results for MNLI when varying the student's hidden size (without initialization of student layers, 4 hidden layers) in Figure~\ref{fig:mnl_hs_app}.

\subsection{Varying the Initialization}
We provide the additional results when varying the initialization for MNLI in Figure~\ref{fig:mnl_init}.

\subsection{Semantic Textual Similarity}
Finally, we also provide the results of our distillation on STS in Figure~\ref{fig:sts}.

\end{document}